\documentclass[preprints,article,accept,information,moreauthors,pdftex]{Definitions/mdpi} 

\firstpage{1} 
\pubvolume{xx}
\issuenum{1}
\articlenumber{5}
\pubyear{2020}
\copyrightyear{2020}
\history{Received: date; Accepted: date; Published: date}
\Title{Looking back to lower-level information in few-shot learning}

\Author{Zhongjie Yu $^{1}$ and Sebastian Raschka $^{1,*\dagger}$}

\AuthorNames{Zhongjie Yu, Sebastian Raschka}

\address{%
$^{1}$ \quad University of Wisconsin-Madison, Department of Statistics}

\corres{Correspondence: sraschka@wisc.edu}

\firstnote{Current address: 1300 University Ave, Medical Sciences Building, Madison, WI 53706, USA} 
\abstract{Humans are capable of learning new concepts from small numbers of examples. In~contrast, supervised deep learning models usually lack the ability to extract reliable predictive rules from limited data scenarios when attempting to classify new examples. This challenging scenario is commonly known as few-shot learning. Few-shot learning has garnered increased attention in recent years due to its significance for many real-world problems. Recently, new methods relying on meta-learning paradigms combined with graph-based structures, which model the relationship between examples, have shown promising results on a variety of few-shot classification tasks. However, existing work on few-shot learning is only focused on the feature embeddings produced by the last layer of the neural network. {The~novel contribution of this paper is the utilization of lower-level information to improve the meta-learner performance in few-shot learning. In~particular, we propose the Looking-Back method, which could use lower-level information to construct additional graphs for label propagation in limited data settings.} Our experiments on two popular few-shot learning datasets, \textit{mini}ImageNet and \textit{tiered}ImageNet, show that our method can utilize the lower-level information in the network to improve state-of-the-art classification performance.}


\usepackage[absolute,overlay]{textpos}

\begin{document}
%%%%%%%%%%%%%%%%%%%%%%%%%%%%%%%%%%%%%%%%%%

%%%%%%%%%%%%%%%%%%%%%%%%%%%%%%%%%%%%%%%%%%

\begin{textblock*}{18.5cm}(2cm, 1cm) % {block width} (coords) 
   Published in \textit{Information} 2020, 11, 345. DOI: 10.3390/info11070345 
\end{textblock*}

\section{Introduction}
Deep learning (DL) is already ubiquitous in our daily lives, including image-based object detection~\cite{liu2020deep}, face recognition~\cite{wani2020supervised}, medical imaging, and healthcare~\cite{wang2020medical}. While DL is outperforming traditional machine learning methods in these aforementioned application areas~\cite{raschka2020machine}, a major downside of DL is that it requires large amounts of data to achieve good performance~\cite{lecun2015deep}. Few-shot learning (FSL) is a subfield of DL that focuses on training DL models under scarce data regimes, thereby opening possibilities for applying DL to new problem areas where the amount of labeled data is limited. 

{In~FSL settings, datasets are comprised of large numbers of categories (i.e.,~class labels), but~only a few examples per class are available. The~main objective of FSL is the design of methods that achieve good generalization performance from the limited number of examples per category. The~overarching concept of FSL is very general and applies to different data modalities and tasks like image classification~\cite{wang2019generalizing}, object detection~\cite{kang2019few}, and text classification~\cite{bao2019few}.   However, most FSL research is focused on image classification so that we will use the terms \textit{examples} and \textit{images} (in a supervised learning context) interchangeably.}

Most FSL methods use an episodic training strategy known as meta-learning~\cite{vinyals2016matching}, where a meta-learner is trained on (classification) tasks with the goal to learn to perform well on new, unseen tasks.  {Many of the most recent FSL methods are based on episodic meta-learning, such as prototypical networks~\cite{snell2017prototypical}, relation networks~\cite{sung2018learning}, model agnostic meta-learning frameworks~\cite{finn2017model}, and LSTM-based meta-learning~\cite{Ravi2017}. Another successful approach to FSL is the use of transfer learning, where models are trained on large datasets and then appropriately transferred to smaller datasets that contain the novel target classes; examples include weight imprinting~\cite{qi2018low}, dynamic few-shot object recognition with attention modules~\cite{gidaris2018dynamic}, and few-shot image classification by predicting parameters from activation values~\cite{qiao2018few}.}

Apart from recent developments in FSL, many researchers have recently proposed methods for implementing graph neural networks (GNNs) to extend deep learning approaches for graph-structured data. In~this context, graphs are used as data structures for modeling the relationships (edges) between data instances (nodes){, which was first proposed via the graph neural network model~\cite{scarselli2008graph} and extended via graph convolutional networks~\cite{duvenaud2015convolutional}, semi-supervised graph convolutional networks~\cite{kipf2017semi}, graph attention networks~\cite{velivckovic2018graph}, and message passing neural networks~\cite{gilmer2017neural}.} Since FSL methods are centered around modeling relationships between the examples in the support and query datasets, GNNs have also gained a growing interest in FSL research, {including approaches aggregating node information from densely connected support and query image graphs~\cite{Garcia2018}, transductive inference~\cite{Liu2019}, and edge-labeling~\cite{kim2019edge}.} GNNs can be computationally prohibitive on large datasets. However, we shall note that one of the significant characteristics of FSL is that datasets for meta-training and meta-testing contain only ``few'' examples per class, such that the computational cost of graph construction becomes small in FSL. 

Previous research has shown that FSL can be improved by incorporating additional information. {For instance, unlabeled data, which is used in conventional~\cite{ren2018meta}, self-trained~\cite{li2019learning}, and transfer learning-based~\cite{yu2020transmatch} semi-supervised FSL, could improve the predictive performance of FSL models. Also, FSL benefits from the inclusion of additional modalities (e.g., textual information describing the images to be classified), which was demonstrated via an adaptive cross-model approach enhancing metric-based FSL~\cite{xing2019adaptive} as well as cross-modal FSL utilizing latent features from aligned autoencoders~\cite{schonfeld2019generalized}.} While the aforementioned works showed that additional \textit{external} information benefits FSL, we raise the question of whether additional \textit{internal} information can be useful as well. 

While the incorporation of additional information can be beneficial, the~utilization of additional \textit{internal} information is not very common in FSL research, and only two recent research papers explored this approach, {i.e.,~Li et al.'s deep nearest neighbor neural network~\cite{li2019revisiting} and the dense classification network by Lifchitz et al.~\cite{lifchitz2019dense}}. In~these works, the~researchers expanded the feature embeddings {(the~low-dimensional representation)} of the data inputs (i.e.,~images), {extracted} from the last layer in the neural network, to higher-dimensional embeddings. These higher-dimensional embeddings were split into several smaller vectors, such that multiple embedding vectors correspond{ed} to the same image. 

In~the DN4 model proposed by Li et al.~\cite{li2019revisiting}, the~last layer's feature embeddings were expanded to form many local descriptors. The~dense classification network by Lifchitz et al.~\cite{lifchitz2019dense} expanded the feature embeddings to three separate vectors that are used for computing the cross-entropy loss during~training.

When it comes to utilizing additional internal information, both DN4~\cite{li2019revisiting} and the dense classification network~\cite{lifchitz2019dense} only considered the last layer's information. In~contrast to existing work on FSL, we consider additional information that is hidden in the earlier layers of the neural network. We hypothesize that such internal information benefits an FSL model's predictive performance. More~specifically, the~extra information hidden in the network considered in this work is comprised of the feature embeddings that can be obtained from layers before the last layer. We propose using a graph structure to integrate this lower-level information into the neural network, {since graph structures are well-suited for modeling relationships in data.}

We refer to the FSL method proposed in this paper as \textit{Looking-Back}, because unlike DN4~\cite{li2019revisiting} and the dense classification network~\cite{lifchitz2019dense}, {this method fully utilizes previous layers' feature embeddings (i.e.,~lower-level information)} rather than focusing on the final layer's feature embeddings alone. During training, the~lower-level information is expected to help the meta-learner to absorb more information overall. Although this lower-level information may not be as useful as the embedding vectors obtained from the last layer, we hypothesize that the lower-level information has a positive impact on the meta-learner. To test this hypothesis, we adopt the widely used Conv-64F~\cite{li2019revisiting} in few-shot learning as a backbone, and {construct graphs for label propagation, following the transductive propagation network (TPN)~\cite{Liu2019}, to capture lower-level information.}

Besides the feature embeddings of the last layer, the~previous layers' feature embeddings are also used for computing the pair-wise similarities between the inputs, based on relational network modules~\cite{Liu2019}. In~the Looking-Back method, three groups of pair-wise similarity measures are computed. The~similarity scores between all support and query images in one episode amount to three separate graph Laplacians, which are used for iterative label propagation, to generate three separate cross-entropy losses. As the experimental results indicate, the~losses from lower-level features are used during meta-training to enhance the performance of the meta-learner. After meta-training, we adopt the last layer's feature embeddings for testing on new tasks (i.e.,~images with class labels that are not seen during training) in a transductive fashion. As the experimental results reveal, the~resulting FSL models have a better predictive performance on new, unseen tasks compared to models generated by meta-learners that 
{do not} utilize lower-level information. 

The~contributions of this work can be summarized as follows:

\begin{enumerate}
    \item We propose a {novel FSL meta-learning method}, Looking-Back,  that utilizes lower-level information from hidden layers, which is different from existing FSL methods that only use feature embedding of the last layer during meta-training.
    \item We implement our Looking-Back method using a graph neural network, which fully utilizes the advantage of graph structures for few-shot learning to absorb the lower-level information in the hidden layers of the neural network.
    \item We evaluate our proposed Looking-Back method on two popular FSL datasets, \textit{mini}ImageNet and \textit{tiered}ImageNet, and achieve new state-of-the-art results, providing supporting evidence that using lower-level information could result in better meta-learners in FSL tasks.
\end{enumerate}

\section{Related Work}

In~this section, we discuss the recent developments in FSL with a focus on methods related to our work. We group these related FSL methods into two main categories, meta-learning-based approaches and transfer learning-based approaches.

\subsection{Meta-Learning}
FSL, based on meta-learning, typically uses episodic training strategies. In~each episode, the~meta-learner is trained on a meta-task, which can be thought of as an image classification task. During training, these tasks are drawn randomly from the training dataset across the episodes. During the model evaluation, tasks are chosen from a separate test dataset, which consists of images from novel classes that are not contained in the training dataset.  

In~$N$-way-$k$-shot FSL, when a meta-learner is trained on several tasks sampled from the training dataset, each training task is subdivided into a support set and a query set. Each task consists of $N$ unique class labels, and the support set consists of $k$ labeled images per class. Utilizing the support set, the~model learns to predict the image labels in the query set. After training, the~meta-learner is then evaluated on new tasks sampled from the test set. Similar to the training tasks, each new task consists of $N$ unique class labels with $k$ images (in the support set) each. However, to assess how well the meta-learner performs on new tasks, the~classes in the test dataset are not overlapping with the classes in the training set. 

Based on the general FSL meta-learning framework described above, we can divide meta-learning approaches further into metric-, optimization-, and graph-based meta-learning, which we discuss in the following subsections.

\subsubsection{Metric-Based Meta-Learning}

Metric-based methods are primarily focused on learning feature embeddings that enable similarity comparisons between support and query images. The~Prototypical Network~\cite{snell2017prototypical} used a Euclidean distance measure to compare the feature embeddings of the query images with centroids of the support images in different classes. The~Relation Network~\cite{sung2018learning} constructed an additional network to compute the similarity score between images directly, instead of using the Euclidean distance measure on the images' feature embeddings similar to the Prototypical Network. DN4~\cite{li2019revisiting} used a cosine similarity measure on multiple local descriptors, obtained by expanding the feature embeddings of the last layer to higher dimensions, to find the most similar images via nearest neighbor search.

\subsubsection{Optimization-Based Meta-Learning}

Optimization-based methods are focused on parameter optimization and how to rapidly learn knowledge from limited training images that can be adapted to novel images. The~model agnostic meta-learning framework (MAML)~\cite{finn2017model} learned a general model that can be efficiently fine-tuned to perform well on other tasks using conventional gradient descent-based optimization. While MAML used second-order partial derivatives to train the general model before task-specific fine-tuning, Reptile~\cite{nichol2018first} was a first-order approximation of MAML that simplified the training procedure and boosted computational performance. Ravi and Larochelle~\cite{Ravi2017} introduced a related yet different approach to optimization-based meta-learning. They proposed the use of an LSTM to model the sequence corresponding to the sequential optimization of the model parameters across different tasks. 

\subsubsection{Graph-Based Meta-Learning}

Graph-based meta-learning uses graph structures to model the relationship between query and support images based on relative similarity measures, where each labeled and unlabeled image represents a node in the graph. There are very few treatments of graph-based methods for FSL in the literature; however, the~topic has recently gained more attention in the FSL research community. 

In~2017, Garcia and Bruna~\cite{Garcia2018} proposed the use of a GNN for aggregating node information in an iterative fashion via a message-passing model, where the support and query images are densely connected in the graph. The~edge-labeling graph neural network (EGNN) modified this approach, using edge- rather than node-label information, combined with inter-cluster dissimilarity and intra-cluster similarity measures~\cite{kim2019edge}. Like GNN, the~Transductive Propagation Network (TPN) considered the graph nodes for representing the feature embeddings of the images~\cite{Liu2019}. However, instead of performing inductive inference (that is, predicting test images one by one), TPN used transductive inference to predict the labels of the entire test set at once, which alleviated the low-data problem in FSL and achieved state-of-the-art performance~\cite{Liu2019}.

\subsection{Transfer Learning}

In~contrast to meta-learning, transfer learning is based on a more conventional supervised learning approach. Here, a model is pre-trained on a large dataset with an abundant number of examples per class. After pre-training on these base classes, the~model is then transferred (i.e.,~fine-tuned) to the novel classes in a few-shot task.

The~weight imprinting~\cite{qi2018low} method constructed classifiers for novel tasks by imprinting the centroids of the novel images' feature embeddings on classifier weights. TransMatch~\cite{yu2020transmatch} extended this concept to semi-supervised settings. The~dynamic few-shot object recognition system proposed by Gidaris and Komodakis introduced an attention module during training to learn the classifier weights~\cite{gidaris2018dynamic}. The~dense classification network was another method based on imprinting~\cite{lifchitz2019dense}. In~addition, this method expanded the feature embeddings, {obtained} from the last layer, to a set of vectors when computing the cross-entropy loss during training on the base classes in the training. All~cross-entropy loss terms were aggregated to compute the overall loss during backpropagation.

\begin{figure}[H]
    \centering
    \includegraphics[width=1.0\textwidth]{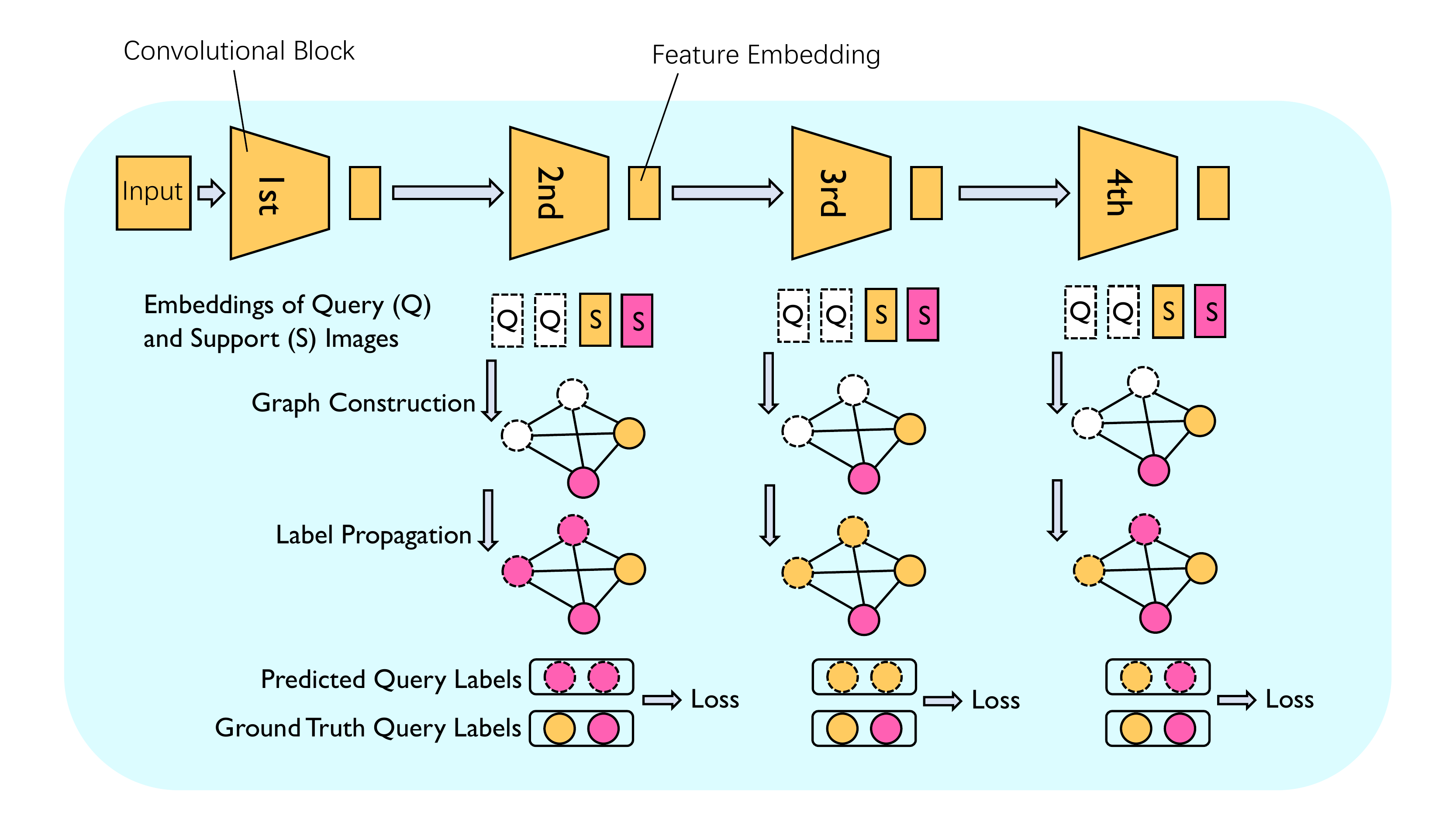}
    \caption{A conceptual overview of the proposed Looking-Back method.}
    \label{fig:proposed_framework}
\end{figure}

The~Looking-Back method we propose in this paper (Figure~\ref{fig:proposed_framework}) uses the same graph construction approach as TPN~\cite{Liu2019}. However, Looking-Back incorporates the feature embeddings from hidden layers in the graph construction procedure as well. We shall note that the simultaneous training with graphs built on lower-level information could also be seen as a particular case of multi-task learning or incremental learning, which was mentioned in~\cite{mallya2018packnet} but is rarely adopted in FSL.

\section{Proposed Method}

In~this section, we introduce our proposed Looking-Back approach utilizing lower-level information to enhance the predictive performance of FSL models. 

\subsection{Problem Definition}

The~goal of FSL is to train predictive models that learn from and perform well on classification tasks, given only a few labeled examples per class. For instance, $N$-way $K$-shot classification can be understood as a classification task with {$N$} unique classes, where {$K$} labeled examples per class are provided for supervised learning.

In~an $N$-way $K$-shot setting, the~dataset for a given task is divided into a support set $\mathbf{S} = \{ (\mathcal{X}_s, \mathcal{Y}_s)\}$ and a query set $\mathbf{Q} = \{ (\mathcal{X}_q, \mathcal{Y}_q)\}$. $\mathbf{S}$ consists of $N \times K$ examples $\mathcal{X} = (\mathbf{x}_1, \mathbf{x}_2, \dots, \mathbf{x}_{N \times K})$ and the corresponding class labels $\mathcal{Y} = (y_1, y_2, \dots, y_{N\times K})$. The~goal is to utilize $\mathbf{S}$ to predict the class labels $(y{'}_1, y{'}_2, \dots, y{'}_{Q \times K})$ for the $Q \times K$ examples in $\mathbf{Q}$, $(\mathbf{x}{'}_1, \mathbf{x}{'}_2, \dots, \mathbf{x}{'}_{Q \times K})$.

Given a large training dataset $D_{base}$, with base classes $\mathbf{C}_{base}$, FSL meta-learning approaches sample many different $N$-way $K$-shot classification tasks $\mathcal{T}=\{T_1, T_2, ..., T_m\}$ randomly from $D_{base}$, to train the meta-learner for $m$ episodes. After training, the~meta-learner is given a novel $N$-way $K$-shot classification task {$T_{novel}$}, such that the {$N$} classes do not overlap with the base classes in $D_{base}$ encountered during training. The~dataset corresponding to {$T_{novel}$} is split into support and query sets, and the meta-learner uses the $N \times K$ labeled examples in the support set to classify the $Q \times K$ examples in the query set.

A successful FSL meta-learner learns from the training tasks $T_1, T_2, ..., T_m$ how to efficiently utilize the few labeled examples in the support set of a novel task {$T_{novel}$} so that the resulting model is able to predict the class labels in the unlabeled query set with good generalization performance.

Considering the general problem definition of FSL and meta-learning given above, the~examples in the query set can be used in a transductive manner as suggested by~\cite{Liu2019}, i.e.,  instead of classifying the query examples one at a time, the~whole query set can be propagated into the network all at once, which improves the predictive performance compared to classifying each query example independently~\cite{Liu2019}.

\subsection{Feature Extractor Module}
\label{sec:feature-extractor}

The~two predominant types of neural network backbone architectures used in FSL research are ResNet-12~\cite{mishra2017simple,oreshkin2018tadam,lee2019meta,sun2019meta} and Conv-64F~ \cite{vinyals2016matching,snell2017prototypical,sung2018learning,Garcia2018,li2019revisiting,Liu2019}. In~this work, we adopt Conv-64F since it is easier to experiment with. However, we shall note that our proposed method is architecture-agnostic and can be implemented for other types of feedforward neural networks.

Conv-64F contains four convolutional blocks where every block is constructed by one convolutional layer with 64 filters of size 3~$\times$~3, followed by a batch normalization layer, ReLU activation, and a 2~$\times$~2 max-pooling layer. Both the convolutional layers and the max-pooling layers have a stride of 1.

Besides extracting feature embeddings from the last layer of the last convolutional block, the~proposed Looking-Back also extracts the embeddings from the last layer of the second and third convolutional block. These three feature embeddings are then used in the graph-based label propagation, as illustrated in Figure~\ref{fig:proposed_framework}. The~dimensions of the feature embeddings extracted by the three convolutional blocks are 64~$\times$~21~$\times$~21, 64~$\times$~10~$\times$~10, and 64~$\times$~5~$\times$~5, respectively. Here, the~number of channels, 64, is determined by the Conv-64F architecture, whereas the channel heights and widths are a consequence of the input image dimensions given the Conv-64F architecture.

\begin{figure}[H]
    \centering
    \includegraphics[width=1.0\textwidth]{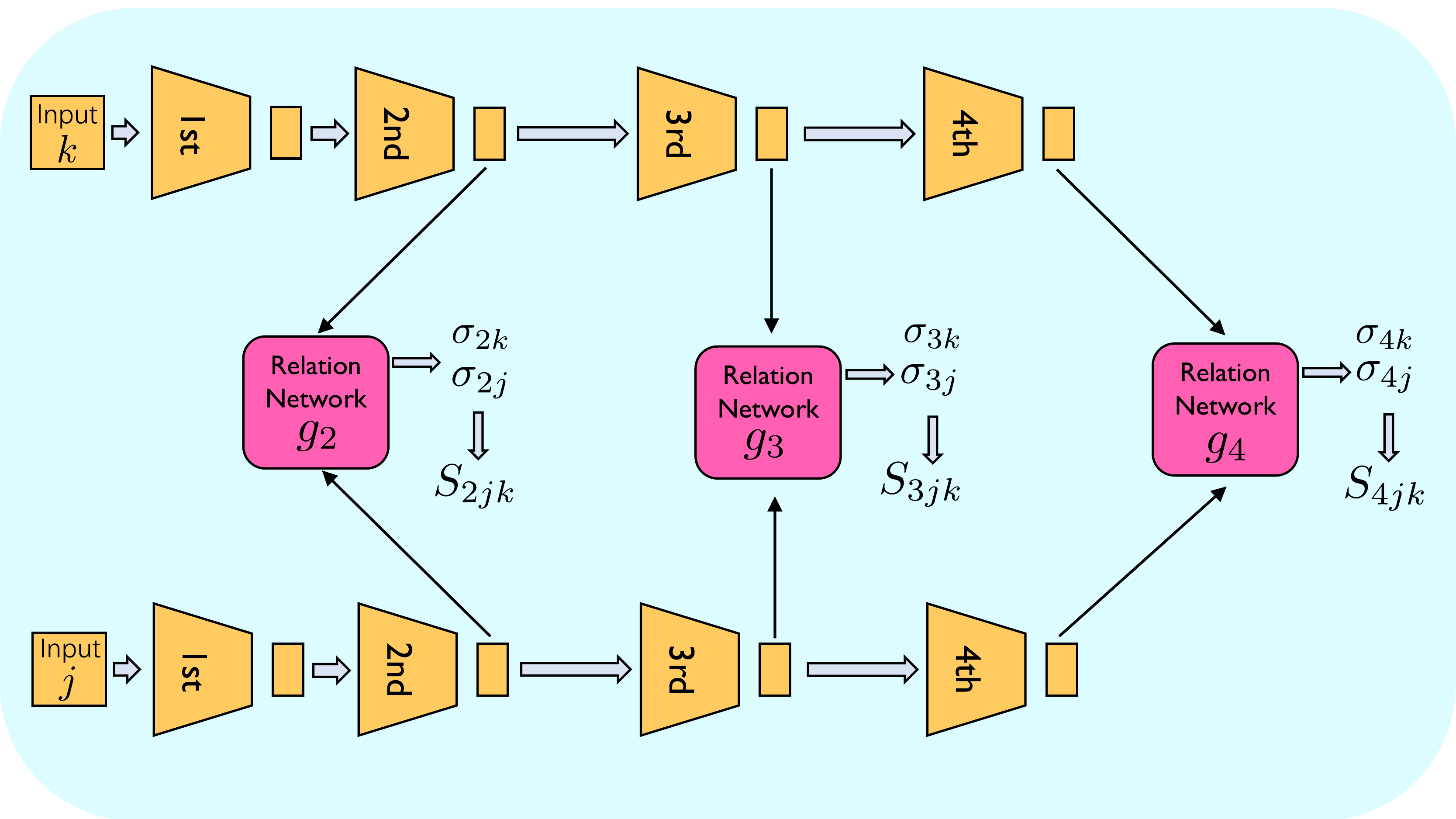}
    \caption{Computing the similarity between a pair of images, inputs $j$ and $k$, based on the feature embeddings produced by the 2nd, 3rd, and 4th convolutional block (layer). The~similarity values computed by the relation network modules are then used to construct multiple graphs for label~propagation.}
    \label{fig:similarity}
\end{figure}

\subsection{Graph Construction Module}
\label{lab:graph-constrution}

In~the original work of TPN~\cite{Liu2019}, the~authors proposed a pair-wise similarity function that used an example-wise length-scale parameter. Adopting this mechanism, for the output of $i$-th convolutional block, we compute the similarity of two images ($j,k$) via
\begin{equation}
S_{ijk}=\exp\left( -\frac{1}{2} \left\Vert \frac{f_i(\mathbf{x}_{ij})}{\sigma_{ij}} - \frac{f_i(\mathbf{x}_{ik})}{\sigma_{ik}} \right\Vert_2^2 \right),
\end{equation}
which measures the distance between the two feature embeddings. Here $\sigma_{ij}$ is {a scale parameter for the feature embedding} computed by a relation network module, {which is described in the next paragraph}. As illustrated in Figure~\ref{fig:similarity}, we use a separate relation network for the second, third, and fourth convolutional block, since the dimensions and information contents of the respective feature embeddings differ.

The~overall architecture of the relation network module, which computes $\sigma_{ij}$ and $\sigma_{ik}$, is similar to the architecture used by Liu et al.~\cite{Liu2019}. Each relation network module consists of two convolutional blocks, followed by two fully-connected layers. Each convolutional block is composed of a 3~$\times$~3 convolutional layer with a stride of 1, a batch normalization layer, ReLU activation, and a 2~$\times$~2 max-pooling layer with a stride of 1.

In~the Looking-Back model, we compute multiple symmetric normalized graph Laplacians~\mbox{\cite{chung1997spectral} via} 
\begin{equation}
L_i=D_i^{-1/2}S_{i}D_i^{-1/2},
\end{equation}

{where $D_i$ is the diagonal matrix whose $d$-th diagonal element is the sum of the $d$-th row of the $S_i$. Here, we only keep $m$-max values from every row in $S_i$ to construct a $m$-nearest neighbor graph for each layer during episodic training to improve computational efficiency as suggested by Liu et al.~\cite{Liu2019}.}

\subsection{Classification Loss}

After constructing {different}  graphs {for multiple layers} as explained in Section~\ref{lab:graph-constrution}, label propagation~\cite{zhou2004learning} is used to compute the prediction (i.e.,~class-membership) scores for the query images~\cite{Liu2019}.

Let $P^{(0)}$ be an initial score matrix. For a given image $\langle \mathbf{x}_j, y_j \rangle$ in the support set, 
\begin{equation}P^{(0)}_{jl}=\left\{\begin{array}{ll}
0 & \text { if } y_j \neq l, \\
1 & \text { if } y_j = l.
\end{array}\right.
\end{equation}

The~label propagation process is an iterative process
\begin{equation}
\label{eq:prop-1}
P^{(t+1)}=\alpha L_i P^{(t)} +(1-\alpha) P^{(0)},
\end{equation}
where $P^{(t)}$ is the predicted label at time step $t$. The~predicted scores $P^*_i$ for an input image's feature embedding from the $i$-th convolutional block are computed via 
\begin{equation}
\label{eq:prop-2}
P^*_i=(I-\alpha L_i)^{-1}P^{(0)},
\end{equation}
where $I$ is the identity matrix, $L_i$ is the normalized graph Laplacian of that feature embedding from the $i$-th convolutional block, and $\alpha$ is a hyperparameter controlling propagation rate.
  
After computing the prediction scores, we obtain class-membership probability scores for the feature embeddings from the $i$-th convolutional block by applying a softmax function as follows:
\begin{equation}
p(\hat{y}_{ij}=k|\mathbf{x}_{ij})=\frac{\exp(P^*_{i,jk})}{\sum_{k=1}^N \exp(P^*_{i,jk})},
\end{equation}
where $\hat{y}_{ij}$ is the predicted class label for feature embedding of the $j$-th input image from the $i$-th convolutional block, and $P^*_{i,jk}$ is the predicted score at the $k$-th position.

The~total loss term is the combination of cross-entropy loss for different layers' features:
\begin{equation}
\label{eq:loss}
\text{Loss}=-\sum_i \sum_{j=1}^{N\times K+Q} \sum_{k=1}^N w_i \text{I}(y_{ij}=k) \log(p(\hat{y}_{ij}=k|\mathbf{x}_{ij})),
\end{equation}
where $w_i$ is a relative weight for the cross-entropy loss term of the feature embeddings from the $i$-th convolutional block and is a hyperparameter during the episodic training.
 
The~feature embeddings from the second ($i=2$) and third ($i=3$) convolutional block containing lower-level information are only used during training to improve the feature extractor module (Section~\ref{sec:feature-extractor}). In~both the validation and test stage, the~class labels $\hat{\mathbf{y}}=\underset{k}{\arg \max} \; p(\hat{\mathbf{y}}_{i}=k|\mathbf{x}_{i})$ are obtained from the prediction on feature embeddings of the last convolutional block only, that is, the~fourth convolutional block, $i=4$.

\section{Experiments} 
In~this section, we evaluate the proposed Looking-Back method on two popular FSL benchmark datasets, i.e.,~\textit{mini}ImageNet~\cite{Ravi2017} and \textit{tiered}ImageNet~\cite{ren2018meta}, and compare with other state-of-the-art FSL~methods.

\subsection{Datasets}
\label{sec:datasets}
{\bf \textit{{mini}}ImageNet.} 

The~\textit{mini}ImageNet dataset is widely used for comparing different few-shot learning methods~\cite{Ravi2017}. It is a small subset of ImageNet~\cite{deng2009imagenet} that consists of 100 classes with 600 examples per class. For our experiments, we split the dataset into 64 classes for training, 16 classes for validation, and 20 classes for testing following~\cite{Ravi2017}. 

{\bf \textit{tiered}ImageNet.}
Similar to \textit{mini}ImageNet, the~\textit{tiered}ImageNet dataset is a small, simplified version of ImageNet proposed by~\cite{ren2018meta}.
Different from \textit{mini}ImageNet, \textit{tiered}ImageNet has a hierarchical or \textit{tiered} structure consisting of 34 larger classes, where each larger class contains 10 to 30 smaller classes (i.e.,~related subcategories). \textit{tiered}ImageNet contains 608 smaller classes and 779,165 images in total. We split the dataset as described in~\cite{ren2018meta}, resulting in a training set consisting of 20 larger classes, a validation set consisting of 6 larger classes, and the test set consisting of 8 larger classes. The~advantage of splitting the dataset based on the larger classes, as opposed to splitting into the subclasses, is that this approach creates a clearer distinction between training, test, and validation sets.

\subsection{Implementation Details}
\label{sec:implementation-details}
As mentioned before, we adopted the Conv-64F architecture (Section~\ref{sec:feature-extractor}) as the backbone for our model. During training, we used the three layers' feature embeddings as shown in Figures~\ref{fig:proposed_framework} and \ref{fig:similarity}. For~label propagation, we chose the same hyperparameters as described in~\cite{Liu2019}, setting $\alpha$ (the~propagation coefficient, Equations~\eqref{eq:prop-1}~and~\eqref{eq:prop-2}) to 0.99 and $m$ (the per-row max values of the graph Laplacians) to 20. Moreover, we gave equal weighting to the individual loss terms when computing the total loss Equation~\eqref{eq:loss}, that is, setting $w_2$, $w_3$, and $w_4$ to 1.

During the episodic training, each episode was a 5-way $K$-shot task with $15$-query images in each task, mimicking the testing scenario. We used the Adam optimizer~\cite{kingma2014adam} to train the model and set the initial learning rate to 0.001. For \textit{mini}ImageNet, the~learning rate was decayed by a multiplicative factor of 0.8 every 5,000 episodes. The~same multiplicative factor was used for decaying the learning rate when training on \textit{tiered}ImageNet, but it was decayed more frequently, every 2000 epochs, due to the larger size and complexity of \textit{tiered}ImageNet.

To evaluate the model on the test set, we randomly sampled 600 $5$-way $K$-shot tasks from an independent test set with $K=1$ and $K=5$, respectively. In~both scenarios, $K=1$ and $K=5$, there were $15$ query samples in each class (that is, $75$ query examples in total), which were used to compute the prediction accuracy for a given task or episode. To compute the overall prediction accuracy of a given model, we randomly sampled the test set $600$ times and calculated the accuracy by averaging the prediction accuracy across these $600$ episodes.

\subsection{Results and Discussion}

\subsubsection{Overall performance}

 In~this section, we compare our proposed Looking-Back method to other state-of-the-art FSL methods. All~neural network implementations are based on a Conv-64F backbone architecture for feature extraction as described in Section~\ref{sec:feature-extractor}. Following the established conventions, we consider both 5-way 1-shot and 5-way 5-shot settings for the performance comparisons, using the two common FSL benchmark datasets \textit{mini}ImageNet and \textit{tiered}ImageNet as described in Section~\ref{sec:datasets}.  The~accuracy is computed as the average of 600 test episodes (as described in Section~\ref{sec:implementation-details}) with a 95\% confidence interval. As the results for \textit{mini}ImageNet (Table~\ref{tab:result_miniimagenet}) and \textit{tiered}ImageNet (Table~\ref{tab:result_tieredimagenet}) indicate, our proposed Looking-Back method achieves state-of-the-art results on both datasets, in both the 5-way 1-shot and 5-way 5-shot scenarios.

\begin{table}[H]
    \centering
    \caption{Accuracy (in \%) on \textit{mini}ImageNet with 95\% confidence interval. Best results are shown in~bold.}
    \begin{tabular}{cccc}
    \toprule
        \textbf{Method} & \textbf{Extract. Net.} & \textbf{1-Shot }& \textbf{5-Shot}\\
        \midrule
        Matching Net~\cite{vinyals2016matching} \hfill & Conv-64 & 43.56 $\pm$ 0.84 & 55.31 $\pm$ 0.73\\
        Prototypical Net~\cite{Fort2017} \hfill & Conv-64 &49.42 $\pm$ 0.78   &  68.20 $\pm$ 0.66       \\
        Relation Net~\cite{sung2018learning} \hfill & Conv-64 & 50.44 $\pm$ 0.82 & 65.32 $\pm$ 0.70\\
        Reptile~\cite{nichol2018first} \hfill & Conv-64 & 49.97 $\pm$  0.32 & 65.99 $\pm$ 0.58  \\
        GNN~\cite{Garcia2018} \hfill & Conv-64 & 49.02 $\pm$ 0.98  & 63.50 $\pm$ 0.84 \\
        MAML~\cite{finn2017model} \hfill & Conv-64 &  48.70 $\pm$ 1.84  & 63.11 $\pm$ 0.92\\        
        TPN~\cite{Liu2019} \hfill &  Conv-64 &  53.75 $\pm$ 0.86 &  69.43 $\pm$ 0.67\\
   
        \midrule
        Looking-Back & Conv-64 &   \textbf{ 55.91 $\pm$ 0.86} &  \textbf{ 70.99 $\pm$ 0.68}\\
        \bottomrule
    \end{tabular}
    \label{tab:result_miniimagenet}
\end{table}
\unskip
\begin{table}[H]
    \centering
    \caption{Accuracy (in \%) on \textit{tiered}ImageNet with 95\% confidence interval. Best results are shown in~bold.}
    \begin{tabular}{cccc}
    \toprule
        \textbf{Method} & \textbf{Extract. Net.} & \textbf{1-Shot} & \textbf{5-Shot}\\
        \midrule
        Prototypical Net~\cite{Fort2017} \hfill & Conv-64 & 53.31 $\pm$ 0.89   &  72.69 $\pm$ 0.74       \\
        Relation Net~\cite{sung2018learning} \hfill & Conv-64 & 54.48 $\pm$ 0.93 & 71.31 $\pm$ 0.78\\
        Reptile~\cite{nichol2018first} \hfill & Conv-64 & 52.36 $\pm$ 0.23 & 71.03 $\pm$ 0.22\\
        MAML~\cite{finn2017model} \hfill & Conv-64 &  51.67 $\pm$ 1.81 & 70.30 $\pm$ 1.75\\
        TPN~\cite{Liu2019} \hfill &  Conv-64 &  57.53 $\pm$ 0.96 &  72.85 $\pm$ 0.74\\
   
        \midrule
        Looking-Back & Conv-64 &   \textbf{ 58.97 $\pm$ 0.97} &  \textbf{ 73.59 $\pm$ 0.74}\\
        \bottomrule
    \end{tabular}
    \label{tab:result_tieredimagenet}
\end{table}

\subsubsection{Comparing Looking-Back and TPN training in a ``Higher Shot'' setting} 
The~performance comparisons between Looking-Back and TPN~\cite{Liu2019} (Tables~\ref{tab:result_miniimagenet} and~\ref{tab:result_tieredimagenet}) provides supportive evidence that utilizing lower-level information, which is contained in previous layers' feature embeddings, improves the predictive performance by a substantial amount by our {Looking-Back method}. In~this section, we investigate whether the lower-level information can also enhance the performance in a ``Higher Shot''~setting.

In~FSL, it is common to use support sets of similar size during meta-training and testing. However, some researchers found that using larger support sets during meta-training (i.e.,~increasing the number of ``shots'') can improve the predictive performance of FSL systems based on evaluation on the same (i.e.,~smaller shot) test sets~\cite{snell2017prototypical,li2019revisiting}. Similar observations have been made in the original TPN paper~\cite{Liu2019}, where the authors described that increasing the number of examples in the support sets during meta-training (referred to as ``Higher Shot'') can improve the predictive accuracy during testing. However, using a larger number of shots during meta-training than testing does not always improve the predictive performance, and it is still an open area of research~\cite{cao2019theoretical}.

Although ``Higher Shot'' training is not the focus of this paper, we conducted experiments with higher shots and report the results in Table~\ref{tab:higher_shots}, adopting the procedure described in the original TPN paper~\cite{Liu2019} to enable fair comparisons. The~results in Table~\ref{tab:higher_shots} indicate that Looking-Back utilizing lower-level information outperforms TPN in a ``Higher Shot'' setting as well.
\begin{table}[H]
    \centering
    \caption{Accuracy (in \%) after training with higher shots. Here, the~system evaluated on a 1-shot test set was trained in a 5-shot setting, and the system evaluated on a 5-shot test was trained in a 10-shot setting. Best results are shown in bold.}
    \begin{tabular}{cccc}
    \toprule
    \textbf{Dataset} &  \textbf{Method} & \textbf{1-Shot} & \textbf{5-Shot}\\ 
    \midrule
      \multirow{2}*{\textit{mini}ImageNet} & TPN & 55.51 $\pm$ 0.86 & 69.86 $\pm$  0.65  \\
     ~ & Looking-Back & \textbf{56.49 $\pm$ 0.83}   & \textbf{70.47 $\pm$ 0.66 }   \\
     \midrule
     \multirow{2}*{\textit{tiered}ImageNet} & TPN & 59.91 $\pm$ 0.94  & 73.30 $\pm$ 0.75  \\ 
     ~ & Looking-Back & \textbf{61.19 $\pm$ 0.92} & \textbf{73.78 $\pm$ 0.74} \\
     \bottomrule
    \end{tabular}
    \label{tab:higher_shots}
\end{table}

Table~\ref{tab:gain from lower-level} summarizes the performance gain of Looking-Back over TPN for the regular meta-training scenario (same number of shots in the training and test tasks, Tables~\ref{tab:result_miniimagenet} and~\ref{tab:result_tieredimagenet}) and meta-training with higher shots (Table~\ref{tab:higher_shots}). From Table~\ref{tab:gain from lower-level}, we can observe that on both datasets, the~improvement of \textit{same} versus \textit{higher} shot meta-training in 1-shot settings is more significant than in 5-shot settings. We argue that when more support images are available (higher shot), the~role of utilizing lower-level information becomes less important. The~main rationale behind using previous layers' feature embeddings is to use additional lower-level information when information from the final layer's feature embedding is scarce. Intuitively, the~role of using lower-level information degrades if a meta-learner can utilize a larger number of examples in the support set.

\begin{table}[H]
    \caption{Performance gain (in \% points) of Looking-Back vs transductive propagation network (TPN) on the test sets when using lower-level information in same-shot (Same) and higher-shot (Higher) training.}
    \centering
    \begin{tabular}{cccc}
    \toprule
    \textbf{Training Approach} &  \textbf{Dataset} &\textbf{ 1-Shot }& \textbf{5-Shot} \\ 
    \midrule
      \multirow{2}*{Same} & \textit{mini}ImageNet & 2.16 & 1.56 \\
     ~ & \textit{tiered}ImageNet & 1.44  & 0.74 \\
     \midrule
     \multirow{2}*{Higher} & \textit{mini}ImageNet & 0.98  &   0.61\\ 
     ~ & \textit{tiered}ImageNet & 1.28  & 0.48  \\
     \bottomrule
    \end{tabular}
    \label{tab:gain from lower-level}
\end{table}

\subsubsection{Influence of higher-shot training on Looking-Back} 

As indicated by the results in Table~\ref{tab:gain from lower-level} and hypothesized in the previous section, our Looking-Back method could be more useful when the data is more scarce. This is likely because the more information is available during training (i.e.,~the support sets consist of additional examples in higher-shot settings), the~more negligible the information from earlier layers becomes as supportive information.

 \begin{table}[H]
    \centering
    \caption{Performance gain (in \% points) of Looking-Back when trained with higher shots compared to training with same shots.}
    \begin{tabular}{ccc}
    \toprule
    \textbf{Dataset} &  \textbf{1-Shot} &\textbf{ 5-Shot} \\ 
    \midrule
      \textit{mini}ImageNet & 0.58 & $-$0.52 \\
   %   ine
      \textit{tiered}ImageNet & 2.22  & 0.19 \\
     \bottomrule
    \end{tabular}
    \label{tab: gain from training with higher shots}
\end{table}

In~a 1-shot setting, we were still able to observe that the lower-level information used by Looking-Back models benefits the model performance when training in the higher shots setting, as summarized in Table~\ref{tab: gain from training with higher shots}. However, in the presence of a larger number of images, using lower-level information during training results in more limited improvements (5-shot test setting on \textit{tiered}ImageNet) or may have a small detrimental impact (5-shot test settings on \textit{mini}ImageNet) as shown in Table~\ref{tab: gain from training with higher shots}. This finding provides further evidence that the lower-level information has a more beneficial effect when the data is more scarce.

\subsubsection{Why only using the last layer's information during inference}
 
Both DN4~\cite{li2019revisiting} and the dense classification network~\cite{lifchitz2019dense} use the entire expanded feature embeddings of the last layer during training as well as inference. One of the main reasons we only use the feature embeddings of the last layer during inference is that the lower-level information from previous layers is used to augment the graph construction during training but does not have equal relevance for the prediction task during inference. In~contrast to Looking-Back, in both DN4 and the dense classification network, the~additional information of the expanded feature embeddings are on the same footing.

To test our hypothesis that the feature embeddings of the last layer bear the most relevance for the prediction task, we compared the prediction accuracy of Looking-Back when using different layers for the class label prediction. As indicated by the results in Table~\ref{tab:different layer}, the~prediction accuracy of the 4th (last) layer is higher than the prediction accuracy of the 3rd layer, and the accuracy of the 3rd layer is higher than the accuracy of the 2nd layer, supporting the hypothesis that the last layer contains the most useful information.

\begin{table}[H]
    \centering
    \caption{Different layers' prediction accuracy (in \%) on 5-way tasks after label propagation with same-shot training.}
    \begin{tabular}{ccccc}
    \toprule
    \textbf{Dataset} &  \textbf{Setting} &\textbf{ 2nd Layer} & \textbf{3rd Layer} & \textbf{4th Layer }\\ 
    \midrule
      \multirow{2}*{\textit{mini}ImageNet} & 1-shot & 42.24 $\pm$ 0.76 & 50.87 $\pm$ 0.81  & 55.91 $\pm$ 0.86  \\
     ~ & 5-shot & 58.10 $\pm$ 0.72 & 67.07 $\pm$ 0.69  & 70.99 $\pm$ 0.68\\
     \midrule
     \multirow{2}*{\textit{tiered}ImageNet} & 1-shot & 46.25 $\pm$ 0.87  & 54.70 $\pm$ 0.93 & 58.97  $\pm$ 0.97  \\ 
     ~ & 5-shot & 61.12 $\pm$ 0.75  & 69.94 $\pm$ 0.74 & 73.59 $\pm$ 0.74   \\
     \bottomrule
    \end{tabular}
    \label{tab:different layer}
\end{table}

\section{Conclusions}

In~this paper, we propose a new approach to FSL {that captures} additional information inside the feature extracting network to improve prediction performance. In~particular, the~proposed Looking-Back {method} employs a graphical structure to utilize the lower-level information from previous layers' feature embeddings, which differs from existing methods that only focus on expansions of the last layer's feature embeddings. Experiments on two popular FSL datasets provide evidence that the utilization of lower-level information in FSL improves the performance of FSL~meta-learners.

\vspace{6pt}

%%%%%%%%%%%%%%%%%%%%%%%%%%%%%%%%%%%%%%%%%%
\authorcontributions{{Conceptualization, Z.Y. and S.R.; investigation and validation, S.R. and Z.Y.; data curation, Z.Y.; writing---original draft preparation, Z.Y. and S.R.; writing---review and editing, Z.Y. and S.R.; visualization, Z.Y. and S.R.; supervision, S.R.; project administration, S.R.; funding acquisition, S.R. All~authors have read and agreed to the published version of the manuscript.}}

\funding{Support for this review article was provided by the Office of the Vice Chancellor for Research and Graduate Education at the University of Wisconsin-Madison with funding from the Wisconsin Alumni Research~Foundation.}

%%%%%%%%%%%%%%%%%%%%%%%%%%%%%%%%%%%%%%%%%%
\conflictsofinterest{The authors declare no conflict of interest.}

%%%%%%%%%%%%%%%%%%%%%%%%%%%%%%%%%%%%%%%%%%
\reftitle{References}

\bibliography{references}{}

\begin{thebibliography}{-------}
\providecommand{\natexlab}[1]{#1}

\bibitem[Liu \em{et~al.}(2020)Liu, Ouyang, Wang, Fieguth, Chen, Liu, and
  Pietik{\"a}inen]{liu2020deep}
Liu, L.; Ouyang, W.; Wang, X.; Fieguth, P.; Chen, J.; Liu, X.; Pietik{\"a}inen,
  M.
\newblock Deep learning for generic object detection: A survey.
\newblock {\em International Journal of Computer Vision} {\bf 2020}, {\em
  128},~261--318.

\bibitem[Wani \em{et~al.}(2020)Wani, Bhat, Afzal, and Khan]{wani2020supervised}
Wani, M.A.; Bhat, F.A.; Afzal, S.; Khan, A.I.
\newblock Supervised deep learning in face recognition. In {\em Advances in
  Deep Learning}; Springer,  2020; pp. 95--110.

\bibitem[Wang \em{et~al.}(2020)Wang, Liang, Chen, Iwamoto, Han, Zhang, Hu, Lin,
  and Chen]{wang2020medical}
Wang, W.; Liang, D.; Chen, Q.; Iwamoto, Y.; Han, X.H.; Zhang, Q.; Hu, H.; Lin,
  L.; Chen, Y.W.
\newblock Medical image classification using deep learning. In {\em Deep
  Learning in Healthcare}; Springer,  2020; pp. 33--51.

\bibitem[Raschka \em{et~al.}(2020)Raschka, Patterson, and
  Nolet]{raschka2020machine}
Raschka, S.; Patterson, J.; Nolet, C.
\newblock Machine learning in Python: Main developments and technology trends
  in data science, machine learning, and artificial intelligence.
\newblock {\em Information} {\bf 2020}, {\em 11},~193.

\bibitem[LeCun \em{et~al.}(2015)LeCun, Bengio, and Hinton]{lecun2015deep}
LeCun, Y.; Bengio, Y.; Hinton, G.
\newblock Deep learning.
\newblock {\em Nature} {\bf 2015}, {\em 521},~436--444.

\bibitem[Wang \em{et~al.}(2019)Wang, Yao, Kwok, and Ni]{wang2019generalizing}
Wang, Y.; Yao, Q.; Kwok, J.T.; Ni, L.M.
\newblock Generalizing from a few examples: A survey on few-shot learning.
\newblock {\em ACM Computing Surveys} {\bf 2019}.

\bibitem[Kang \em{et~al.}(2019)Kang, Liu, Wang, Yu, Feng, and
  Darrell]{kang2019few}
Kang, B.; Liu, Z.; Wang, X.; Yu, F.; Feng, J.; Darrell, T.
\newblock Few-shot object detection via feature reweighting.
\newblock  Proceedings of the IEEE International Conference on Computer Vision,
   2019, pp. 8420--8429.

\bibitem[Bao \em{et~al.}(2019)Bao, Wu, Chang, and Barzilay]{bao2019few}
Bao, Y.; Wu, M.; Chang, S.; Barzilay, R.
\newblock Few-shot Text Classification with Distributional Signatures.
\newblock  International Conference on Learning Representations,  2019.

\bibitem[Vinyals \em{et~al.}(2016)Vinyals, Blundell, Lillicrap, Wierstra,
  et~al.]{vinyals2016matching}
Vinyals, O.; Blundell, C.; Lillicrap, T.; Wierstra, D.; others.
\newblock Matching networks for one shot learning.
\newblock  Advances in Neural Information Processing Systems,  2016, pp.
  3630--3638.

\bibitem[Snell \em{et~al.}(2017)Snell, Swersky, and
  Zemel]{snell2017prototypical}
Snell, J.; Swersky, K.; Zemel, R.
\newblock Prototypical networks for few-shot learning.
\newblock  Advances in Neural Information Processing Systems,  2017, pp.
  4077--4087.

\bibitem[Sung \em{et~al.}(2018)Sung, Yang, Zhang, Xiang, Torr, and
  Hospedales]{sung2018learning}
Sung, F.; Yang, Y.; Zhang, L.; Xiang, T.; Torr, P.H.S.; Hospedales, T.M.
\newblock Learning to compare: relation network for few-shot learning.
\newblock {\em 2018 IEEE/CVF Conference on Computer Vision and Pattern
  Recognition} {\bf 2018}, pp. 1199--1208.

\bibitem[Finn \em{et~al.}(2017)Finn, Abbeel, and Levine]{finn2017model}
Finn, C.; Abbeel, P.; Levine, S.
\newblock Model-agnostic meta-learning for fast adaptation of deep networks.
\newblock  International Conference on Machine Learning,  2017, pp. 1126--1135.

\bibitem[Ravi and Larochelle(2017)]{Ravi2017}
Ravi, S.; Larochelle, H.
\newblock Optimization as a model for few-shot learning.
\newblock  International Conference on Learning Representations,  2017.

\bibitem[Qi \em{et~al.}(2018)Qi, Brown, and Lowe]{qi2018low}
Qi, H.; Brown, M.; Lowe, D.G.
\newblock Low-shot learning with imprinted weights.
\newblock {\em 2018 IEEE/CVF Conference on Computer Vision and Pattern
  Recognition} {\bf 2018}, pp. 5822--5830.

\bibitem[Gidaris and Komodakis(2018)]{gidaris2018dynamic}
Gidaris, S.; Komodakis, N.
\newblock Dynamic few-shot visual learning without forgetting.
\newblock {\em 2018 IEEE/CVF Conference on Computer Vision and Pattern
  Recognition} {\bf 2018}, pp. 4367--4375.

\bibitem[Qiao \em{et~al.}(2018)Qiao, Liu, Shen, and Yuille]{qiao2018few}
Qiao, S.; Liu, C.; Shen, W.; Yuille, A.L.
\newblock Few-shot image recognition by predicting parameters from activations.
\newblock {\em 2018 IEEE/CVF Conference on Computer Vision and Pattern
  Recognition} {\bf 2018}, pp. 7229--7238.

\bibitem[Scarselli \em{et~al.}(2009)Scarselli, Gori, Tsoi, Hagenbuchner, and
  Monfardini]{scarselli2008graph}
Scarselli, F.; Gori, M.; Tsoi, A.C.; Hagenbuchner, M.; Monfardini, G.
\newblock The graph neural network model.
\newblock {\em IEEE Transactions on Neural Networks} {\bf 2009}, {\em
  20},~61--80.

\bibitem[Duvenaud \em{et~al.}(2015)Duvenaud, Maclaurin, Iparraguirre,
  Bombarell, Hirzel, Aspuru-Guzik, and Adams]{duvenaud2015convolutional}
Duvenaud, D.K.; Maclaurin, D.; Iparraguirre, J.; Bombarell, R.; Hirzel, T.;
  Aspuru-Guzik, A.; Adams, R.P.
\newblock Convolutional networks on graphs for learning molecular fingerprints.
\newblock  Advances in Neural Information Processing Systems,  2015, pp.
  2224--2232.

\bibitem[Kipf and Welling(2017)]{kipf2017semi}
Kipf, T.; Welling, M.
\newblock Semi-supervised classification with graph convolutional networks.
\newblock {\em ArXiv} {\bf 2017}, {\em abs/1609.02907}.

\bibitem[Velickovic \em{et~al.}(2018)Velickovic, Cucurull, Casanova, Romero,
  Li{\`o}, and Bengio]{velivckovic2018graph}
Velickovic, P.; Cucurull, G.; Casanova, A.; Romero, A.; Li{\`o}, P.; Bengio, Y.
\newblock Graph attention networks.
\newblock {\em ArXiv} {\bf 2018}, {\em abs/1710.10903}.

\bibitem[Gilmer \em{et~al.}(2017)Gilmer, Schoenholz, Riley, Vinyals, and
  Dahl]{gilmer2017neural}
Gilmer, J.; Schoenholz, S.S.; Riley, P.F.; Vinyals, O.; Dahl, G.E.
\newblock Neural message passing for quantum chemistry.
\newblock  Proceedings of the 34th International Conference on Machine
  Learning-Volume 70,  2017, pp. 1263--1272.

\bibitem[Garcia and Estrach(2018)]{Garcia2018}
Garcia, V.; Estrach, J.B.
\newblock Few-shot learning with graph neural networks.
\newblock  6th International Conference on Learning Representations,  2018.

\bibitem[Liu \em{et~al.}(2019)Liu, Lee, Park, Kim, Yang, Hwang, and
  Yang]{Liu2019}
Liu, Y.; Lee, J.; Park, M.; Kim, S.; Yang, E.; Hwang, S.J.; Yang, Y.
\newblock Learning to propagate labels: transductive propogation network for
  few-shot learning.
\newblock  International Conference on Learning Representations,  2019.

\bibitem[Kim \em{et~al.}(2019)Kim, Kim, Kim, and Yoo]{kim2019edge}
Kim, J.; Kim, T.; Kim, S.; Yoo, C.D.
\newblock Edge-labeling graph neural network for few-shot Learning.
\newblock {\em 2019 IEEE/CVF Conference on Computer Vision and Pattern
  Recognition} {\bf 2019}, pp. 11--20.

\bibitem[Ren \em{et~al.}(2018)Ren, Triantafillou, Ravi, Snell, Swersky,
  Tenenbaum, Larochelle, and Zemel]{ren2018meta}
Ren, M.; Triantafillou, E.; Ravi, S.; Snell, J.; Swersky, K.; Tenenbaum, J.B.;
  Larochelle, H.; Zemel, R.S.
\newblock Meta-Learning for semi-supervised few-shot classification.
\newblock  International Conference on Learning Representations,  2018.

\bibitem[Li \em{et~al.}(2019)Li, Sun, Liu, Zhou, Zheng, Chua, and
  Schiele]{li2019learning}
Li, X.; Sun, Q.; Liu, Y.; Zhou, Q.; Zheng, S.; Chua, T.S.; Schiele, B.
\newblock Learning to self-train for semi-supervised few-shot classification.
\newblock  Advances in Neural Information Processing Systems,  2019, pp.
  10276--10286.

\bibitem[Yu \em{et~al.}(2020)Yu, Chen, Cheng, and Luo]{yu2020transmatch}
Yu, Z.; Chen, L.; Cheng, Z.; Luo, J.
\newblock TransMatch: A Transfer-Learning Scheme for Semi-Supervised Few-Shot
  Learning.
\newblock  Proceedings of the IEEE/CVF Conference on Computer Vision and
  Pattern Recognition,  2020, pp. 12856--12864.

\bibitem[Xing \em{et~al.}(2019)Xing, Rostamzadeh, Oreshkin, and
  Pinheiro]{xing2019adaptive}
Xing, C.; Rostamzadeh, N.; Oreshkin, B.; Pinheiro, P.O.
\newblock Adaptive cross-modal few-shot learning.
\newblock  Advances in Neural Information Processing Systems,  2019, pp.
  4848--4858.

\bibitem[Schonfeld \em{et~al.}(2019)Schonfeld, Ebrahimi, Sinha, Darrell, and
  Akata]{schonfeld2019generalized}
Schonfeld, E.; Ebrahimi, S.; Sinha, S.; Darrell, T.; Akata, Z.
\newblock Generalized zero-and few-shot learning via aligned variational
  autoencoders.
\newblock  2019 IEEE/CVF Conference on Computer Vision and Pattern Recognition,
   2019, pp. 8247--8255.

\bibitem[Li \em{et~al.}(2019)Li, Wang, Xu, Huo, Gao, and Luo]{li2019revisiting}
Li, W.; Wang, L.; Xu, J.; Huo, J.; Gao, Y.; Luo, J.
\newblock Revisiting local descriptor based image-to-class measure for few-shot
  learning.
\newblock {\em 2019 IEEE/CVF Conference on Computer Vision and Pattern
  Recognition} {\bf 2019}, pp. 7253--7260.

\bibitem[Lifchitz \em{et~al.}(2019)Lifchitz, Avrithis, Picard, and
  Bursuc]{lifchitz2019dense}
Lifchitz, Y.; Avrithis, Y.; Picard, S.; Bursuc, A.
\newblock Dense classification and implanting for few-shot learning.
\newblock {\em 2019 IEEE/CVF Conference on Computer Vision and Pattern
  Recognition} {\bf 2019}, pp. 9250--9259.

\bibitem[Nichol \em{et~al.}(2018)Nichol, Achiam, and Schulman]{nichol2018first}
Nichol, A.; Achiam, J.; Schulman, J.
\newblock On first-order meta-learning algorithms.
\newblock {\em ArXiv} {\bf 2018}, {\em abs/1803.02999}.

\bibitem[Mallya and Lazebnik(2018)]{mallya2018packnet}
Mallya, A.; Lazebnik, S.
\newblock {PackNet}: Adding multiple tasks to a single network by iterative
  pruning.
\newblock {\em 2018 IEEE/CVF Conference on Computer Vision and Pattern
  Recognition} {\bf 2018}, pp. 7765--7773.

\bibitem[Mishra \em{et~al.}(2018)Mishra, Rohaninejad, Chen, and
  Abbeel]{mishra2017simple}
Mishra, N.; Rohaninejad, M.; Chen, X.; Abbeel, P.
\newblock A simple neural attentive meta-learner.
\newblock  International Conference on Learning Representations,  2018.

\bibitem[Oreshkin \em{et~al.}(2018)Oreshkin, L{\'o}pez, and
  Lacoste]{oreshkin2018tadam}
Oreshkin, B.; L{\'o}pez, P.R.; Lacoste, A.
\newblock {TADAM}: Task dependent adaptive metric for improved few-shot
  learning.
\newblock  Advances in Neural Information Processing Systems,  2018, pp.
  721--731.

\bibitem[Lee \em{et~al.}(2019)Lee, Maji, Ravichandran, and Soatto]{lee2019meta}
Lee, K.; Maji, S.; Ravichandran, A.; Soatto, S.
\newblock Meta-learning with differentiable convex optimization.
\newblock  2019 IEEE/CVF Conference on Computer Vision and Pattern Recognition,
   2019, pp. 10657--10665.

\bibitem[Sun \em{et~al.}(2019)Sun, Liu, Chua, and Schiele]{sun2019meta}
Sun, Q.; Liu, Y.; Chua, T.S.; Schiele, B.
\newblock Meta-transfer learning for few-shot learning.
\newblock  2019 IEEE/CVF Conference on Computer Vision and Pattern Recognition,
   2019, pp. 403--412.

\bibitem[Chung and Graham(1997)]{chung1997spectral}
Chung, F.R.; Graham, F.C.
\newblock {\em Spectral graph theory}; American Mathematical Soc.,  1997.

\bibitem[Zhou \em{et~al.}(2004)Zhou, Bousquet, Lal, Weston, and
  Sch{\"o}lkopf]{zhou2004learning}
Zhou, D.; Bousquet, O.; Lal, T.N.; Weston, J.; Sch{\"o}lkopf, B.
\newblock Learning with local and global consistency.
\newblock  Advances in Neural Information Processing Systems,  2004, pp.
  321--328.

\bibitem[Deng \em{et~al.}(2009)Deng, Dong, Socher, Li, Li, and
  Li]{deng2009imagenet}
Deng, J.; Dong, W.; Socher, R.; Li, L.J.; Li, K.; Li, F.F.
\newblock Image{N}et: A large-scale hierarchical image database.
\newblock {\em 2009 IEEE/CVF Conference on Computer Vision and Pattern
  Recognition} {\bf 2009}, pp. 248--255.

\bibitem[Kingma and Ba(2014)]{kingma2014adam}
Kingma, D.P.; Ba, J.
\newblock Adam: A method for stochastic optimization.
\newblock {\em ArXiv} {\bf 2014}, {\em abs/1412.6980}.

\bibitem[Fort(2018)]{Fort2017}
Fort, S.
\newblock Gaussian prototypical networks for few-shot learning on omniglot.
\newblock {\em ArXiv} {\bf 2018}, {\em abs/1708.02735}.

\bibitem[Cao \em{et~al.}(2020)Cao, Law, and Fidler]{cao2019theoretical}
Cao, T.; Law, M.T.; Fidler, S.
\newblock A theoretical analysis of the number of sShots in few-shot learning.
\newblock {\em ArXiv} {\bf 2020}, {\em abs/1909.11722}.

\end{thebibliography}

\end{document}